\documentclass[letterpaper, 10 pt, conference]{ieeeconf}  

\IEEEoverridecommandlockouts                              

\usepackage{graphics} 
\usepackage{epsfig} 
\usepackage{mathptmx} 
\usepackage{times} 
\usepackage{amsmath} 
\usepackage{amssymb}  
\usepackage{silence}
\usepackage{subcaption}
\usepackage{algpseudocode}
\usepackage{algorithmicx}
\usepackage{algorithm}
\usepackage{graphicx}
\usepackage{multicol}
\usepackage{multirow}
\usepackage{censor}
\usepackage[utf8]{inputenc}
\title{\LARGE \bf
Semantic Zone-Based Map Management for Stable AI-Integrated Mobile Robots
}

\author{Huichang Yun and Seungho Yoo$^{*}$%
 \thanks{Dept. of computer Engineering, Pukyong National University, Busan, Korea}%
 \thanks{$^{*}$ Corresponding Author}%
}

\begin{document}

\maketitle
\thispagestyle{empty}
\pagestyle{empty}

\begin{abstract}
Recent advances in large AI models (VLMs and LLMs) and joint use of the 3D dense maps, enable mobile robots to provide more powerful and interactive services grounded in rich spatial context. However, deploying both heavy AI models and dense maps on edge robots is challenging under strict memory budgets. When the memory budget is exceeded, required keyframes may not be loaded in time, which can degrade the stability of position estimation and interfering model performance.
We proposes a semantic zone-based map management approach to stabilize dense-map utilization under memory constraints. We associate keyframes with semantic indoor regions (e.g., rooms and corridors) and keyframe management at the semantic zone level prioritizes spatially relevant map content while respecting memory constraints. This reduces keyframe loading and unloading frequency and memory usage.
We evaluate the proposed approach in large-scale simulated indoor environments and on an NVIDIA Jetson Orin Nano under concurrent SLAM--VLM execution. With Qwen3.5:0.8b, the proposed method improves throughput by 3.3 tokens/s and reduces latency by 21.7\% relative to a geometric map-management strategy. Furthermore, while the geometric strategy suffers from out-of-memory failures and stalled execution under memory pressure, the proposed method eliminates both issues, preserving localization stability and enabling robust VLM operation. These results demonstrate that the proposed approach enables efficient dense map utilization for memory constrained, AI-integrated mobile robots. Code is available at: https://github.com/huichangs/rtabmap/tree/segment

\end{abstract}



\section{INTRODUCTION}

With recent advances in robotics and large-scale Artificial Intelligence (AI) models, mobile robots can provide more detailed services that are closely related to daily life.
%
To deliver such services within a given area, robots require detailed information about the surrounding environment, including persistent spatial memory and accurate localization.
%
To obtain this information, maps are essential, and high-capacity 3D dense maps are commonly used in mobile robot services.
%
Building on this foundation, powerful models and rich indoor context enable spatially grounded services beyond basic navigation.
These capabilities allow robots to understand and interpret language instructions that refer to places and objects (e.g., ``\textit{Go to the table next to the sofa in the hallway}'') and execute them reliably using a consistent 3D map frame.
These capabilities significantly expand what robots can do and broaden their potential applications.

%
However, deploying AI models together with large 3D map resources on robots is challenging because they operate in more resource-constrained environments than workstations or servers where VLMs are commonly used, leading to strict memory limits.
%
Even in conventional robot deployments without AI workloads, it is difficult to load the entire map into memory at once, so keyframes are dynamically loaded and unloaded during operation.
%
Under tight memory constraints, fine-grained keyframe retrieval consumes substantial memory, and continual retrieval for new areas forces frequent memory allocation and cleanup, which increases the frequency of load and unload operations and system overhead.
%
More importantly, when available memory is insufficient, loop closure can fail because the system cannot keep enough keyframes in memory.
%
As a result, loop closure outcomes may become unstable or incorrect.

\begin{figure*}[ht]
\begin{subfigure}{0.62\textwidth}
    \includegraphics[width=\linewidth]{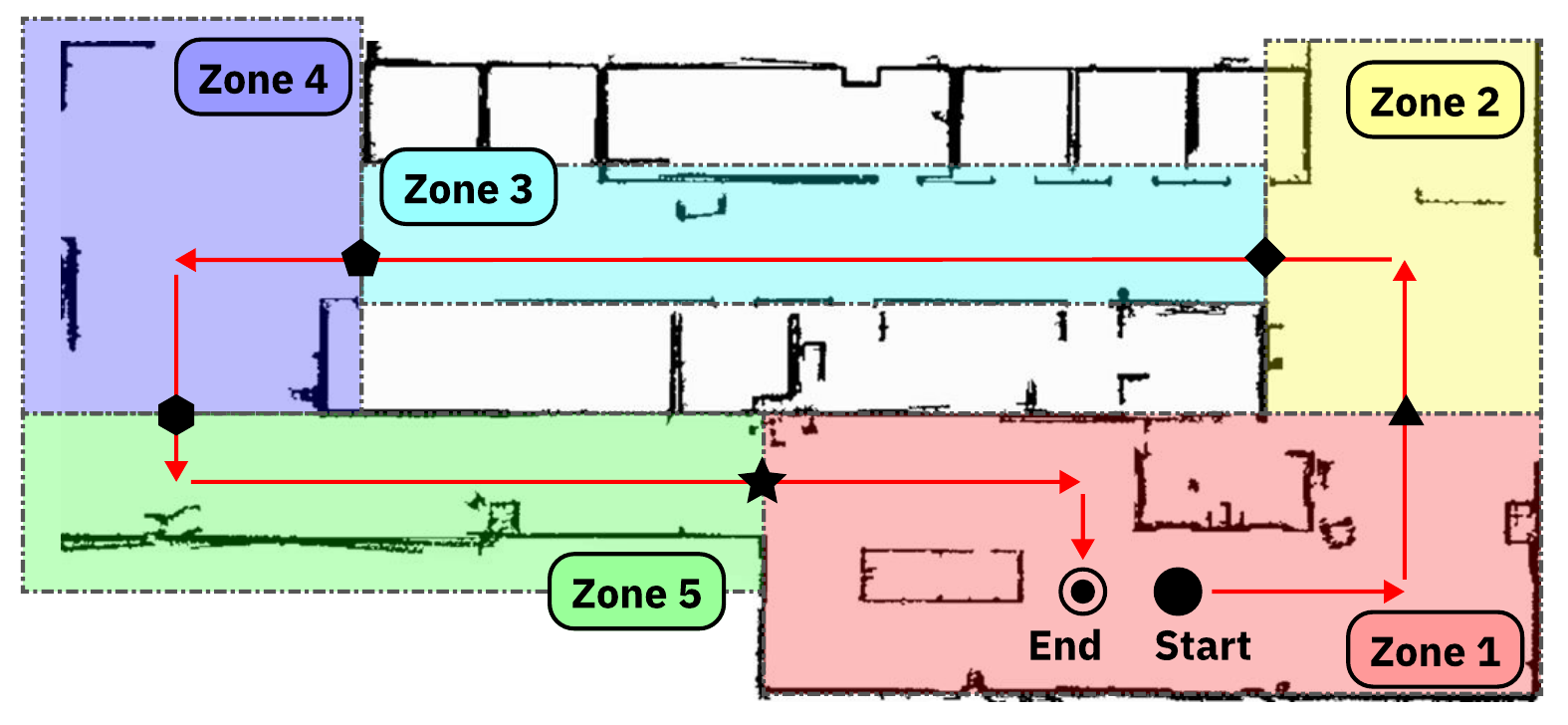}
    \caption{Map of the given envrionment divided into semantic zones with the planned route}
    \label{fig:fig1a}
\end{subfigure}
\hfill
\begin{subfigure}{0.36\textwidth}
    \includegraphics[width=\linewidth]{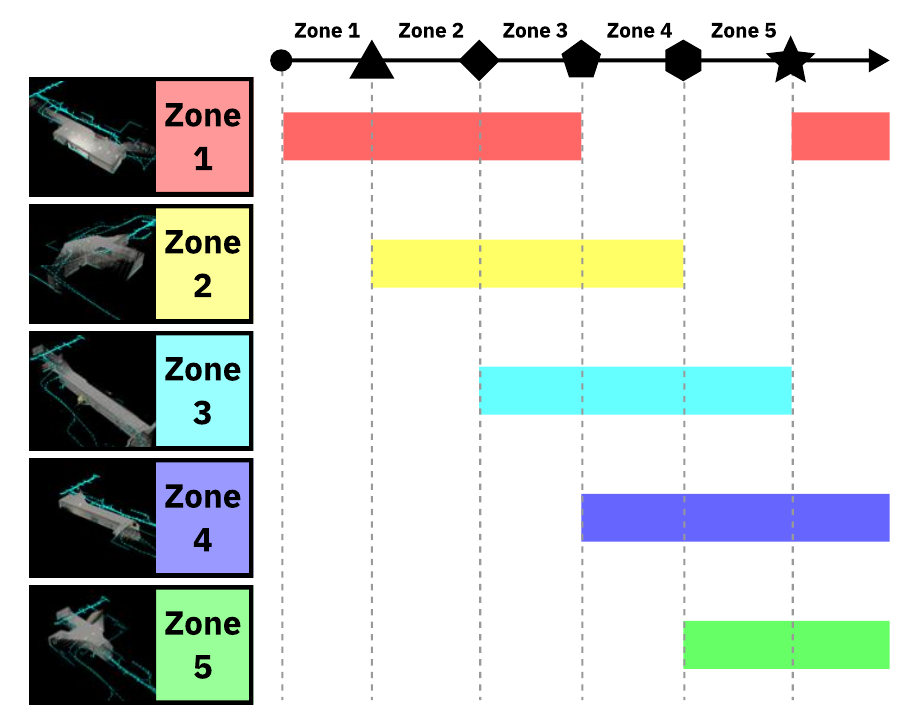}
    \caption{Managed zones for the robot's progress}
    \label{fig:fig1b}
\end{subfigure}
\caption{Overview of the Semantic Zone Based Map Management}
\label{fig:system_overview}
\end{figure*}

To mitigate the issues discussed above, we propose a semantic zone-based map management approach that enables stable 3D dense map utilization under strict memory constraints.
%
Since typical mobile robot motion is continuous in space and time, the system can anticipate which regions are likely to be needed and use this to load map information more effectively.
%
Given the continuity of robot motion, likely travel directions and forthcoming locations can be estimated, allowing keyframes to be prefetched only for zones expected to be relevant.
Also, eviction of keyframes is likewise guided by zone prediction.
%
With this approach, the robot can effectively use its limited memory by combining prediction-based zone eviction with zone-level batch loading and unloading of keyframes, preventing memory saturation.
%
Figure~\ref{fig:system_overview} illustrates the proposed mechanism. Figure~\ref{fig:fig1a} shows the environment partitioned into semantic zones and the robot’s planned route, which provides the basis for predicting likely upcoming zones. Figure~\ref{fig:fig1b} then depicts how only a subset of zones is activated around the robot’s progress, enabling zone-level batch load and unload of keyframes while respecting the memory limit.

%
To realize and validate our proposed approach in large-scale indoor environments, we built on a simulation environment with Nvidia Isaac simulation, and implemented our proposed methods on Jetson Orin Nano\cite{jetsonOrinNano}, which has 8 GB of unified memory, with concurrent execution of VLM models (Gemma\cite{gemma2025} and Qwen\cite{qwen2025}), and RTAB-Map\cite{Labbe2019}.
With the evaluation result, our proposed approach, semantic zone-based keyframe management, could provide a practical mechanism to improve the robustness of dense map utilization on memory-constrained AI-integrated robots.
The contribution of our paper is as follows:

\begin{itemize}
    \item We motivate dense map utilization as a key enabler of spatially grounded, high-quality robot services when combined with large AI models, and identify memory saturation as a practical deployment bottleneck.
    \item We propose a semantic zone-based keyframe load/unload mechanism that strictly enforces a runtime memory limit.
    \item We provide an edge-oriented evaluation with large-scale indoor environments and concurrent VLM execution on Jetson Orin Nano.
\end{itemize}

\section{RELATED WORK}

In this section  we review prior research on memory-efficient map and keyframe management, semantic and topological representations for scalable SLAM, and systems that balance accuracy with resource constraints. And we summarize how these strands motivate our zone-based approach and clarify the gap addressed by prediction-guided memory regulation.

\subsection{Memory-Efficient and Scalable Map/Keyframe Management in SLAM}

Map growth and increasing memory demand are long-standing bottlenecks in large-scale and long-term SLAM. Prior work has pursued memory efficiency either by reducing map size---through pruning, sparsification, or selective keyframe retention---or by preserving map content while constraining only the \emph{resident} working set. Representative examples of the former include information-theoretic pruning for compressing pose graphs \cite{Kretzschmar2011}, sliding-window map sparsification \cite{Zhang2025MSS}, and optimization-based keyframe selection to reduce redundancy under map-size constraints \cite{Thorne2025}. While effective for bounding resources, these approaches may discard map elements that could be useful when the robot revisits areas or when downstream modules require rich spatial context.

Motivated by the growing value of co-executing large AI models (e.g., VLMs) with 3D dense maps for context-grounded robot services, we focus on map-preserving memory management that maintains map quality while limiting in-memory usage. RTAB-Map exemplifies this direction by providing database-backed hierarchical memory and explicit keyframe load/unload mechanisms for long-term operation under bounded resources \cite{Labbe2011,Labbe2013,Labbe2018,Labbe2019}. Complementary scalability techniques further structure maps into higher-level units, such as hierarchical or topological representations and submaps, to localize optimization and restrict search to spatially relevant portions \cite{Estrada2005,Ehlers2020,Aguiar2023,Scucchia2024}. Resource-constrained deployment has also motivated adaptive and hardware-aware SLAM designs that explicitly account for limited compute and memory budgets \cite{Chen2023,SuperNoVA2025}.

However, these works typically decide what to retain or activate using geometric, temporal, or graph-structural cues, and region/submap structures are primarily introduced for scalability of optimization or retrieval rather than for memory-bounded runtime activation in the presence of competing workloads. In contrast, our work applies semantic information directly to map management, partitioning space into semantically meaningful zones to enable more efficient map activation and memory use.

\begin{figure*}[t]
    \centering
    \includegraphics[width=\textwidth]{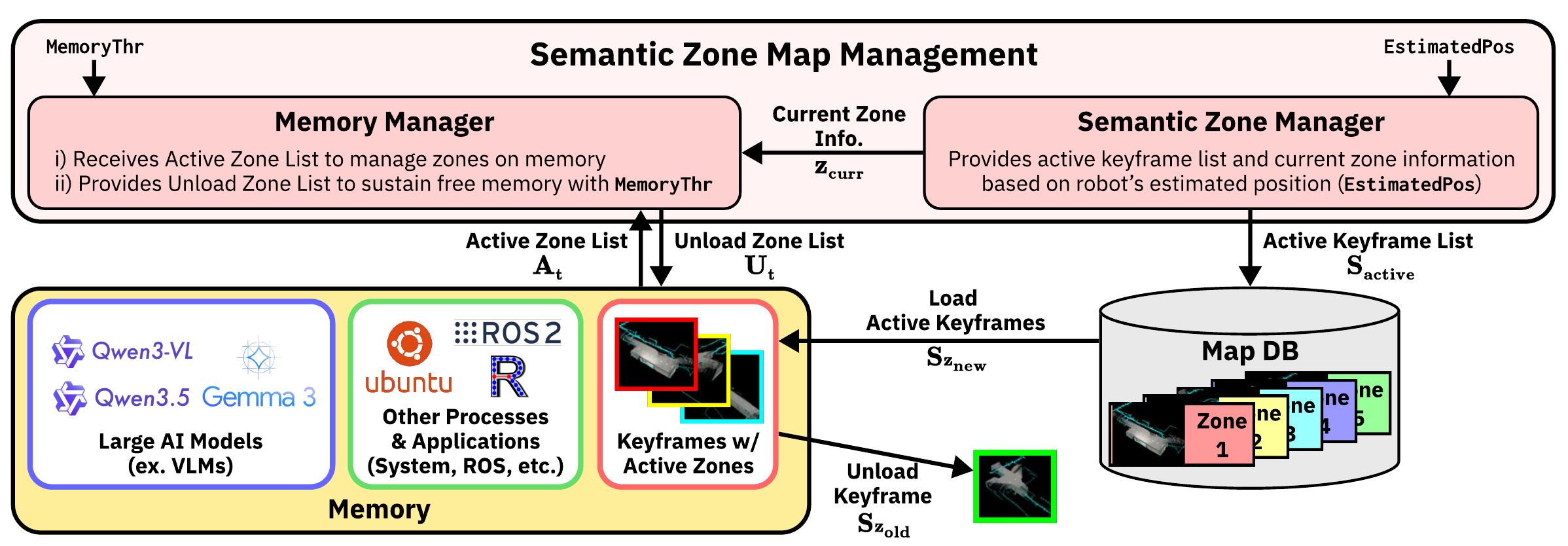}
    \caption{Overview of semantic zone-based keyframe management. The Semantic Zone Manager uses the estimated pose to determine the current zone and to form an active keyframe list. The Memory Manager enforces \texttt{MemoryThr} by maintaining an active zone list and an unload zone list, ensuring strict compliance with the memory limit before loading new keyframes from the map database.}
    \label{fig:keyframe_management_overview}
\end{figure*}


\subsection{Semantic Information for SLAM and Robot Map Utilization}

Semantic information has been widely integrated into SLAM to enhance robustness and scene understanding, including segment-based localization and semantic cues for handling dynamic environments \cite{Dube2020,Cramariuc2021,Wen2021}. Recent surveys highlight a growing emphasis on semantic understanding and open-world robustness, and hierarchical semantic representations have been explored for large-scale reasoning \cite{Miao2025,Bavle2025}. With the rise of large visual models, early systems incorporate foundation-model perception into SLAM pipelines, and recent work investigates VLM-driven keyframe selection to reduce redundancy \cite{Zheng2025,Huo2025}.

Despite this progress, most semantic SLAM efforts focus on perception quality, localization robustness, or map representation. Our work instead leverages semantic zones to regulate runtime keyframe load/unload under memory constraints, improving stability when dense maps and heavy AI models are co-executed on edge platforms.

\section{SEMANTIC ZONE-BASED MAP MANAGEMENT}
Robots equipped with large-scale AI models and 3D dense maps can deliver increasingly diverse and high-quality services. However, in mobile robotic platforms with limited hardware resources, it is not feasible to retain all keyframes stored in the map database in memory simultaneously. As a result, keyframes should be loaded from and unloaded to the map database during runtime.

Conventional systems typically perform load and unload operations based on recency or geometric proximity. Under memory-constrained conditions, however, repeated database access and fine-grained keyframe retrieval can significantly increase the frequency of load and unload operations, imposing substantial overhead on the system. In extreme cases, delayed keyframe loading may destabilize loop closure execution, leading to degraded localization performance.

To address this issue, we propose a semantic zone-based map management approach. Instead of accessing keyframes individually across the entire map, we associate each keyframe with a predefined semantic zone (e.g., rooms, corridors) and perform keyframe management at the zone granularity. By enforcing strict zone-level control and maintaining the number of active keyframes within a predefined memory threshold, the proposed approach stabilizes memory usage and improves the robustness of map utilization.

Figure~\ref{fig:keyframe_management_overview} illustrates an overview of the proposed semantic zone-based map management.

\subsection{Semantic Zones}
We adopt semantic zones as a structural unit for map management. A semantic zone corresponds to a spatially meaningful region, such as a room or corridor. Fig~\ref{fig:fig1a} illustrates an example of semantic zone partitioning in the hospital environment.

The motivation for applying semantic zones arises from the spatial characteristics of service robot operation. Robots typically interact with the environment based on functional regions rather than arbitrary geometric partitions. For instance, when a robot navigates along a corridor, it primarily requires map information associated with that corridor; detailed structure inside adjacent rooms is often irrelevant to the current task, even if those rooms are physically close. This suggests that map utilization naturally follows semantic boundaries.

Maintaining keyframes from spatially or functionally unrelated zones in memory can increase retrieval overhead and memory pressure without contributing to localization or loop closure performance. Organizing keyframes according to semantic zones therefore enables memory management that aligns with the robot’s spatial usage patterns and task relevance.

Formally, we assume that the environment is partitioned into a finite set of semantic zones, $Z = \{z_1, z_2, \ldots, z_K\}$. The proposed method does not depend on how these zones are generated; it only assumes that each keyframe can be associated with one of the predefined regions.

Let $s_i$ denote a keyframe stored in the map database. Each keyframe is assigned to a semantic zone according to its spatial location: $z(s_i) \in Z$. For each zone $z \in Z$, we define the corresponding keyframe set as $S_z = \{s_i \mid z(s_i) = z\}$. This zone-to-keyframe organization will be used in the following subsection to define which keyframes are retained in memory at runtime.

\subsection{Semantic Zone Based Keyframe Management}
Figure~\ref{fig:keyframe_management_overview} illustrates the proposed runtime keyframe management. Given the current estimated pose, the Semantic Zone Manager determines the current semantic zone and updates the set of zones that should be active. Based on this decision, it forms an active keyframe list, which specifies which keyframes should be available for localization and loop closure. The active keyframe list is used to retrieve the corresponding keyframes from the map database.

In parallel, the Memory Manager strictly enforces the memory limit (\texttt{MemoryThr}) by controlling which zones remain loaded in memory. We define the set of zones currently retained in memory as the active zone set,$\ A_t \subseteq Z,$ which corresponds to the \emph{Active Zone List} in Figure~\ref{fig:keyframe_management_overview}. The active keyframes retained in memory are then
\begin{equation}
S_{\text{active}}(t) = \bigcup_{z \in A_t} S_z
\label{eq:active_set}
\end{equation}
Keyframes outside $S_{\text{active}}(t)$ remain stored in the map database and are not kept in memory.

When the Semantic Zone Manager requests activation of a new zone $z_{\text{new}}$, the Memory Manager first predicts the number of keyframes that would become active after the update. Let $K_{\max}$ denote the keyframe capacity implied by MemoryThr, and let $|S_z|$ be the number of keyframes in zone $z$. The predicted active-keyframe count is
\begin{equation}
K_{\text{pred}} = \sum_{z \in A_t} |S_z| + |S_{z_{\text{new}}}|
\label{eq:k_pred}
\end{equation}
If $K_{\text{pred}} > K_{\max}$, the Memory Manager constructs an unload zone set,
$U_t \subseteq A_t,$
which corresponds to the \emph{Unload Zone List} in Figure~\ref{fig:keyframe_management_overview}. Zones in $U_t$ are evicted (e.g., least-recently-used), and all keyframes in those zones are unloaded in batch until the memory constraint is satisfied. Only after this strict eviction step does the system load keyframes of $z_{\text{new}}$ from the map database and updates the active zone set.

This unload-then-load mechanism aligns runtime keyframe availability with semantic zone relevance while strictly respecting the memory limit, thereby stabilizing dense-map localization under shared edge resources. Algorithm 1 shows the detailed runtime procedure of proposed keyframe management.

\begin{algorithm}[t]
\caption{Semantic Zone-Based Keyframe Management with MemoryThr}
\begin{algorithmic}[1]
\State \textbf{Input:} current zone $z_{\mathrm{curr}}$, estimated pose $\hat{\mathbf{x}}_t$, active zone set $A$, keyframe limit \texttt{MemoryThr} ($K_{\max}$)

\State $z_{\mathrm{new}} \leftarrow \textsc{GetZone}(\hat{\mathbf{x}}_t)$
\If{$z_{\mathrm{new}} = \texttt{null}$ \textbf{or} $z_{\mathrm{new}} = z_{\mathrm{curr}}$}
    \State \textbf{return}
\EndIf

\If{$z_{\mathrm{new}} \notin A$}
    \State $K_{\mathrm{pred}} = \sum_{z \in A} |S_z| + |S_{z_{\mathrm{new}}}|$
    
    \While{$K_{\mathrm{pred}} > K_{\max}$}
        \State $z_{\mathrm{old}} \leftarrow \textsc{SelectUnloadZone}(A)$
        \State $A = A \setminus \{z_{\mathrm{old}}\}$
        \State \textsc{UnloadKeyframes}($S_{z_{\mathrm{old}}}$)
        \State $K_{\mathrm{pred}} = K_{\mathrm{pred}} - |S_{z_{\mathrm{old}}}|$
    \EndWhile
    
    \State $A = A \cup \{z_{\mathrm{new}}\}$
    \State \textsc{LoadKeyframes}($S_{z_{\mathrm{new}}}$)
\EndIf

\State $z_{\mathrm{curr}} = z_{\mathrm{new}}$
\end{algorithmic}
\end{algorithm}

\section{IMPLEMENTATION}
We adopt RTAB-Map as the baseline SLAM system for implementing our map-management approach, and we use its default keyframe retrieval and memory-management behavior as the baseline strategy. Many SLAM frameworks control map growth through pruning or culling, or through sliding-window optimization, thereby reducing resident memory by discarding map elements rather than swapping them between long-term storage and a bounded working set. In contrast, RTAB-Map provides a hierarchical memory structure consisting of working, short-term, and long-term memory, with explicit mechanisms to unload keyframes from active memory and reload them from the on-disk database when revisiting previously mapped areas. RTAB-Map’s default load and unload decisions are primarily driven by geometric proximity and pose-graph structure, making it a natural geometric baseline for evaluating the benefit of semantic zone–based map management under an otherwise identical SLAM pipeline. These properties make RTAB-Map well suited for implementing and evaluating our approach.

We integrate the proposed method by modifying RTAB-Map’s memory-management pipeline only. We reuse RTAB-Map’s built-in keyframe load and unload interfaces to manage the in-memory working set, while keeping pose estimation, loop-closure detection, and graph optimization unchanged.

During the offline phase, we assign a semantic zone label to each stored keyframe and persist it as lightweight metadata indexed by the keyframe ID. We maintain a mapping from keyframe ID to zone for fast lookup and, for each zone, a list of keyframe IDs belonging to that zone, enabling zone-granularity load and unload operations.

During online localization, after RTAB-Map updates the pose estimate for each incoming sensor update, we determine the zone corresponding to the current pose. A zone-transition event triggers a memory update. When a zone is activated, we load all keyframes assigned to that zone from the database into active memory. When a zone is evicted, we unload all keyframes in that zone, preserving them in the database while removing them from the in-memory working set. Issuing load and unload at zone granularity reduces the number of database transactions compared to fine-grained per-keyframe retrieval.

We enforce a strict memory budget using a keyframe-capacity threshold, denoted \texttt{MemoryThr}, which specifies the maximum number of resident keyframes. Before activating a new zone, we predict the number of resident keyframes using only zone-level counts from the metadata. If the prediction exceeds \texttt{MemoryThr}, we iteratively evict zones according to a zone-level least-recently-used policy and unload their keyframes until the constraint is satisfied. We then load the requested zone, ensuring that memory usage does not temporarily exceed the budget.

Once loaded, keyframes are consumed by RTAB-Map’s localization and loop-closure modules through the standard processing flow. Thus, our implementation controls which keyframes are resident at runtime, while RTAB-Map’s original procedures, including loop-closure verification and graph optimization, operate normally on the currently loaded subset.

\section{EXPERIMENTS AND RESULTS}
We conducted experiments to evaluate the proposed semantic zone based memory management approach and analyze the impact of utilizing VLM and maps in an edge resource environment.
The objectives of these experiments were to determine how the proposed semantic region-based map management approach improves memory performance during dense map operations and whether these memory improvements impact the performance of AI models such as VLM.
Specifically, we evaluate (i) the reduction in memory footprint and load/unload churn achieved by semantic zone based keyframe management, and (ii) the resulting impact on concurrent SLAM--VLM execution.

\subsection{Test Environments}
Experiments were conducted in large-scale indoor environments built in NVIDIA Isaac Sim, including a hospital scenario. Localization was performed using a pre-built 3D dense map, and online map updates were disabled during evaluation to isolate the effects of runtime map utilization and memory regulation. Semantic zones were manually defined to partition the environment into spatially meaningful indoor regions (e.g., rooms and corridors), and keyframes in the map database were associated with these zones. Unless otherwise noted, \textit{Basic} denotes the default RTAB-Map keyframe loading/unloading mechanism, while \textit{Semantic} refers to our semantic zone-based map management strategy.

All experiments were executed on an NVIDIA Jetson Orin Nano with 8GB unified memory to emulate resource-constrained edge deployment. We attempt \texttt{qwen3:2b/4b, qwen3.5:0.8b/2b, gemma3:4b} as the VLM workload because it represents a deployable edge-scale model while still imposing substantial memory pressure on an 8GB platform. In our setup, the model runs concurrently with SLAM and tends to consume most available memory during inference, making it suitable for stress-testing memory contention.

\begin{table*}[!ht]
\centering
\caption{Impact of concurrent VLM execution on dense-map utilization performance.}
\label{tab:rtabmap_compare_transposed}
\renewcommand{\arraystretch}{1.15}
\begin{tabular}{l|c|c|c|c|c|c|c}
\hline
\multirow{2}{*}{Metric} 
& \multicolumn{3}{c|}{Basic} 
& \multicolumn{3}{c|}{Semantic} 
& \multirow{2}{*}{Basic vs Semantic + VLM (\%)} \\
\cline{2-7}
& Only & +VLM & Impact (\%) 
& Only & +VLM & Impact (\%) 
& \\
\hline
Update Rate (FPS) [Hz] $\uparrow$
& 0.96 & 0.65 & -32
& 0.96 & 0.95 & -1
& +46 \\
Loop Closures $\uparrow$
& 53 & 31 & -42
& 35 & 33 & -6
& +6 \\
Avg. TimeTotal [ms] $\downarrow$
& 779.41 & 5064.87 & +550
& 566.06 & 784.81 & +39
& -85 \\
ATE RMSE [m] $\downarrow$
& 0.14 & 0.31 & +128
& 0.18 & 0.19 & +6
& -39 \\
RPE Translation RMSE [m] $\downarrow$
& 0.20 & 0.42 & +112
& 0.24 & 0.22 & -7
& -48 \\
RPE Rotation RMSE [deg] $\downarrow$
& 0.38 & 1.00 & +163
& 0.57 & 0.32 & -44
& -68 \\
\hline
\end{tabular}
\end{table*}

\subsection{Semantic Zone-based Memory Management Results}

\begin{figure}[t]
    \centering
    \begin{subfigure}{\columnwidth}
        \centering
        \includegraphics[width=\linewidth]{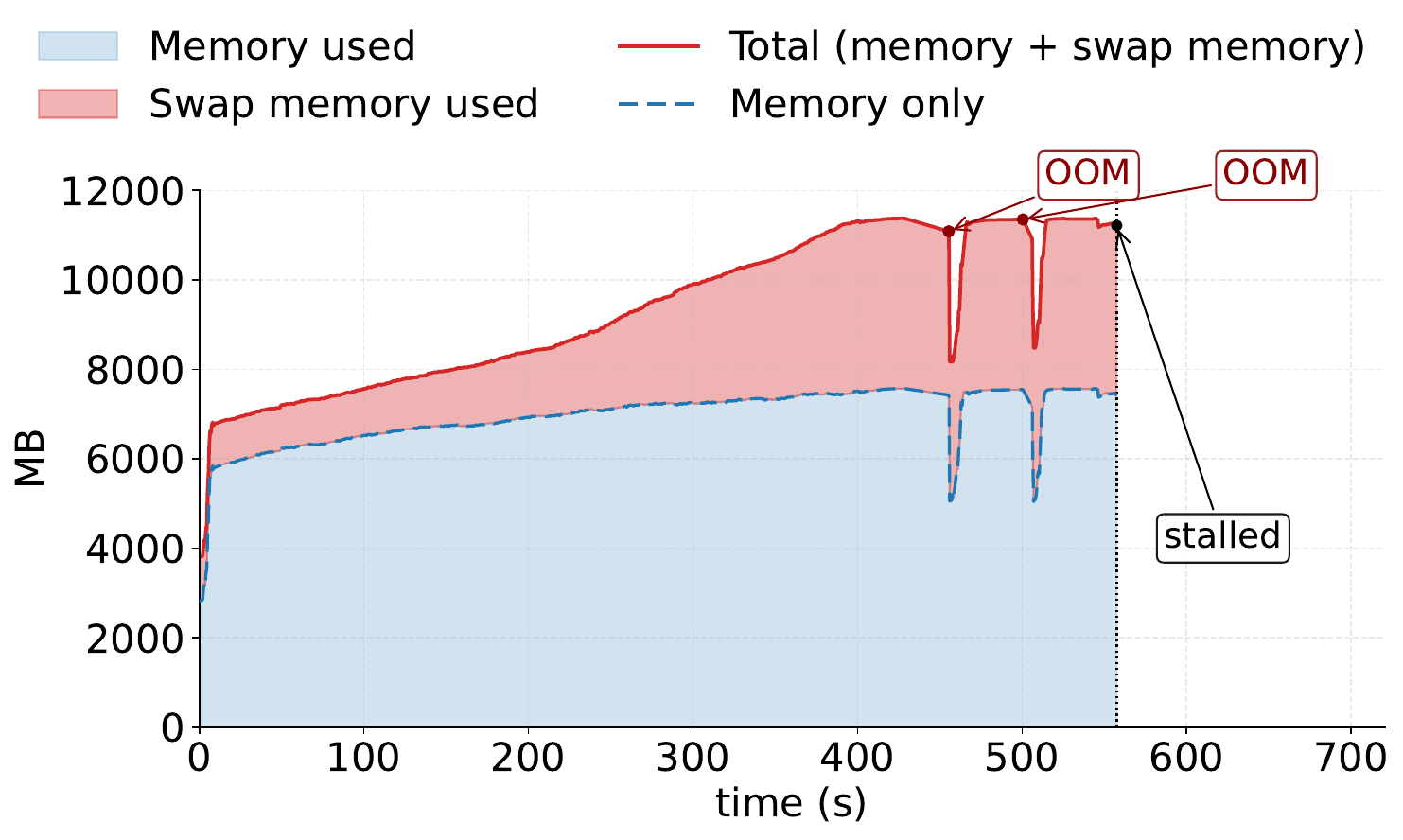}
        \caption{Basic + VLM}
        \label{fig:memory_usage_basic_vlm}
    \end{subfigure}

    \vspace{0.5em}

    \begin{subfigure}{\columnwidth}
        \centering
        \includegraphics[width=\linewidth]{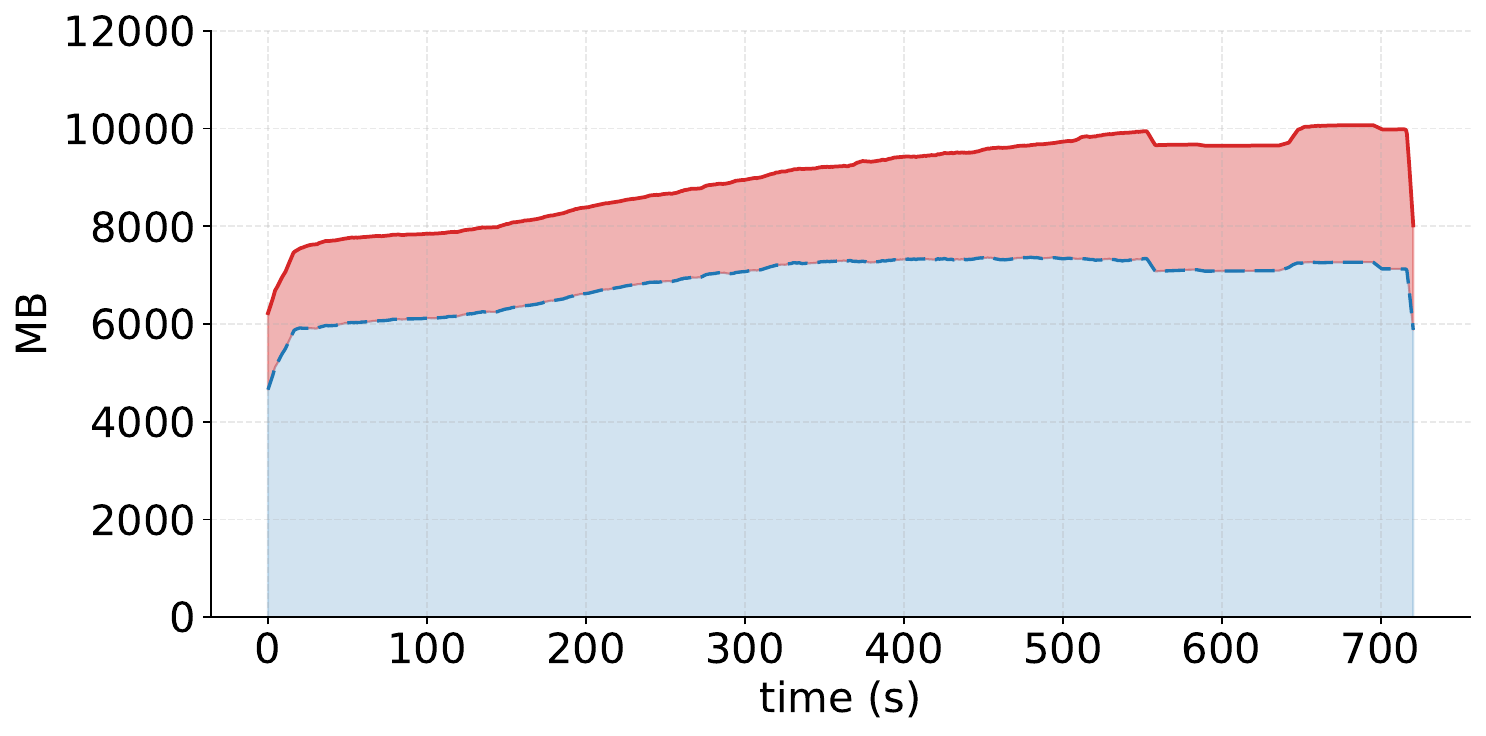}
        \caption{Semantic + VLM}
        \label{fig:memory_usage_semantic_vlm}
    \end{subfigure}

    \caption{Memory usage comparison between (a) Basic + VLM and (b) Semantic + VLM.}
    \label{fig:memory_usage}
\end{figure}

Before analyzing the other experimental results, we first evaluated the effectiveness of the proposed memory management method. The results are presented in figure~\ref{fig:memory_usage}. As shown in figure~\ref{fig:memory_usage}(a), the Basic + VLM configuration repeatedly reached memory saturation, resulting in two out-of-memory (OOM) events and eventual stalled execution before complete whole execution.
In contrast, as shown in figure~\ref{fig:memory_usage}(b), the Semantic + VLM configuration completed the experiment without OOM or stalled behavior and maintained a more stable overall memory usage pattern. These results suggest that semantic zone-level keyframe management alleviates memory contention between SLAM and the VLM, thereby preserving localization continuity and enabling more stable dense-map utilization on resource-constrained edge platforms.

\subsection{Impact of Concurrent SLAM-VLM Execution on SLAM Performance}

\begin{figure}[t]
    \centering
    \includegraphics[width=\linewidth]{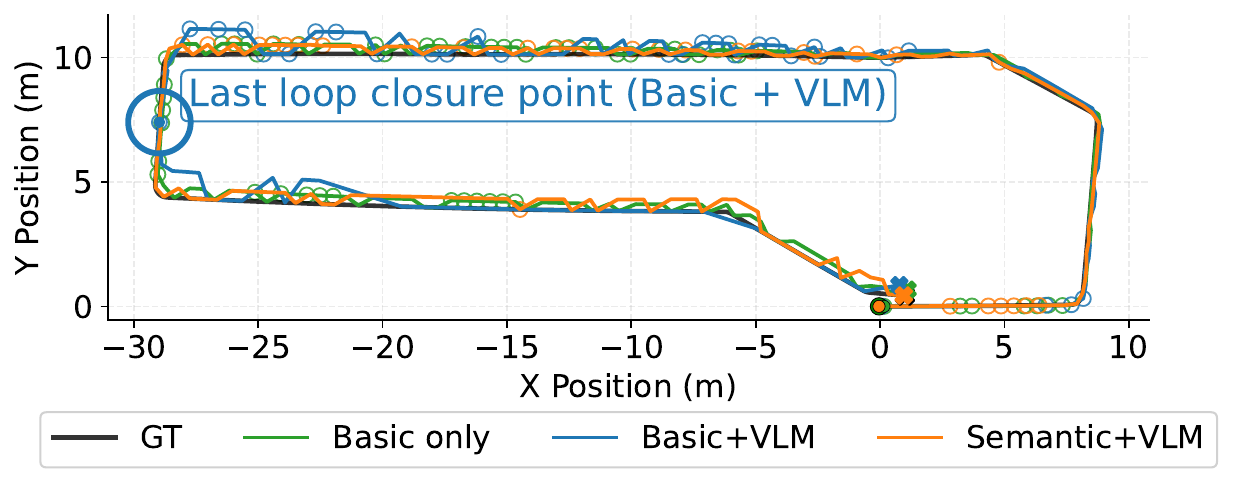}
    \caption{Trajectory comparison among GT, Basic only, Basic+VLM, and Semantic+VLM. The blue circle indicates the final accepted loop-closure position in the Basic+VLM run. After this point, no additional loop closures are accepted until the end of the run.}
    \label{fig:experiment_trajectories}
\end{figure}

Table~\ref{tab:rtabmap_compare_transposed} summarizes the impact of concurrent VLM execution on SLAM performance under the basic and  zone configurations. Overall, the basic configuration exhibits substantial degradation when VLM inference is executed concurrently, whereas the semantic zone memory management approach remains comparatively stable.

In the basic algorithm, enabling VLM reduces the update rate from 0.96~Hz to 0.65~Hz and decreases the number of accepted loop closures from 53 to 31. The mean total processing time also increases sharply from 779.41~ms to 5064.87~ms. This runtime slowdown is accompanied by clear degradation in localization accuracy: ATE RMSE increases from 0.14~m to 0.31~m, translational RPE RMSE increases from 0.20~m to 0.42~m, and rotational RPE RMSE increases from 0.38$^\circ$ to 1.00$^\circ$. These results indicate that the basic memory-management configuration is strongly affected by concurrent VLM execution, which reduces loop-closure opportunities and significantly degrades trajectory quality.

\begin{table*}[t]
\centering
\caption{VLM token generation throughput (TPS) and latency comparison.}
\label{tab:vlm_tps_latency_all}
\renewcommand{\arraystretch}{1.2}
\begin{tabular}{c|c|ccc|ccc|cc}
\hline
\multicolumn{2}{c|}{} & \multicolumn{3}{c|}{Basic} & \multicolumn{3}{c|}{Semantic} & \multicolumn{2}{c}{Relative Change} \\
\cline{1-2} \cline{3-5} \cline{6-8} \cline{9-10}
Model & Size & Execution & Avg. TPS & Avg. Lat. (ms) & Execution & Avg. TPS & Avg. Lat. (ms) & TPS (\%) $\uparrow$ & Lat (\%) $\downarrow$ \\
\hline
Qwen3-vl & 2B (1.9GB) & $\triangle$ & 32.43 & 9642 & \raisebox{-0.2ex}{{\Large$\circ$}} & 33.36 & 10260 & 3.1 & 6.4 \\

 & 4B (3.3GB) & {\Large$\times$} & - & - & {\Large$\times$} & - & - & - & - \\
\hline
Qwen3.5 & 0.8B (1GB) & $\triangle$ & 29.80 & 30396 & \raisebox{-0.2ex}{{\Large$\circ$}} & 31.22 & 23812 & 3.3 & -21.7 \\

 & 2B (2.7GB) & {\Large$\times$} & - & - & {\Large$\times$} & - & - & - & - \\
\hline
Gemma3 & 4B (3.3GB) & \raisebox{-0.2ex}{{\Large$\circ$}} & 14.38 & 13175 & \raisebox{-0.2ex}{{\Large$\circ$}} & 15.19 & 11880 & 7.1 & -9.8 \\
\hline
\multicolumn{10}{l}{\footnotesize Execution legend: \raisebox{-0.2ex}{{\Large$\circ$}} completed the full scenario, $\triangle$ stalled during execution, {\Large$\times$} failed to start.} \\
\end{tabular}
\end{table*}

\begin{figure*}[!t]
    \centering
    \begin{subfigure}[t]{0.24\linewidth}
        \centering
        \includegraphics[width=\linewidth]{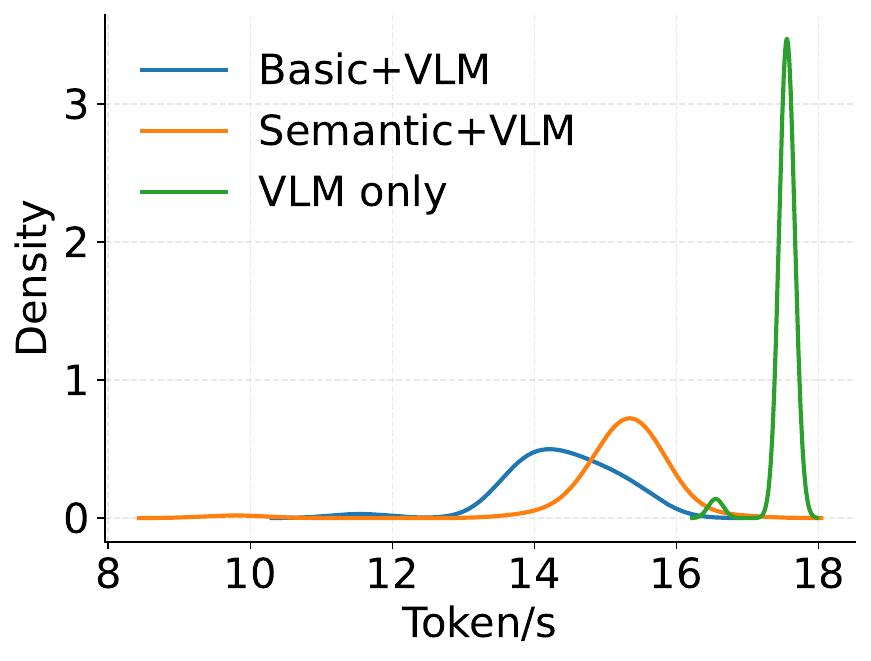}
        \caption{Token/s PDF}
        \label{fig:vlm_token_s_pdf}
    \end{subfigure}
    \begin{subfigure}[t]{0.24\linewidth}
        \centering
        \includegraphics[width=\linewidth]{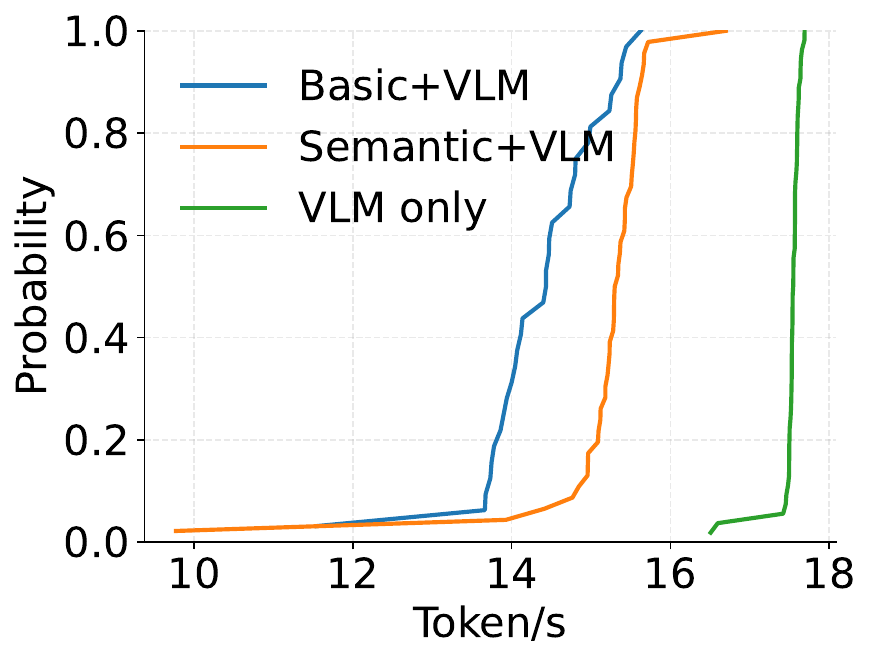}
        \caption{Token/s CDF}
        \label{fig:vlm_token_s_cdf}
    \end{subfigure}
    \begin{subfigure}[t]{0.24\linewidth}
        \centering
        \includegraphics[width=\linewidth]{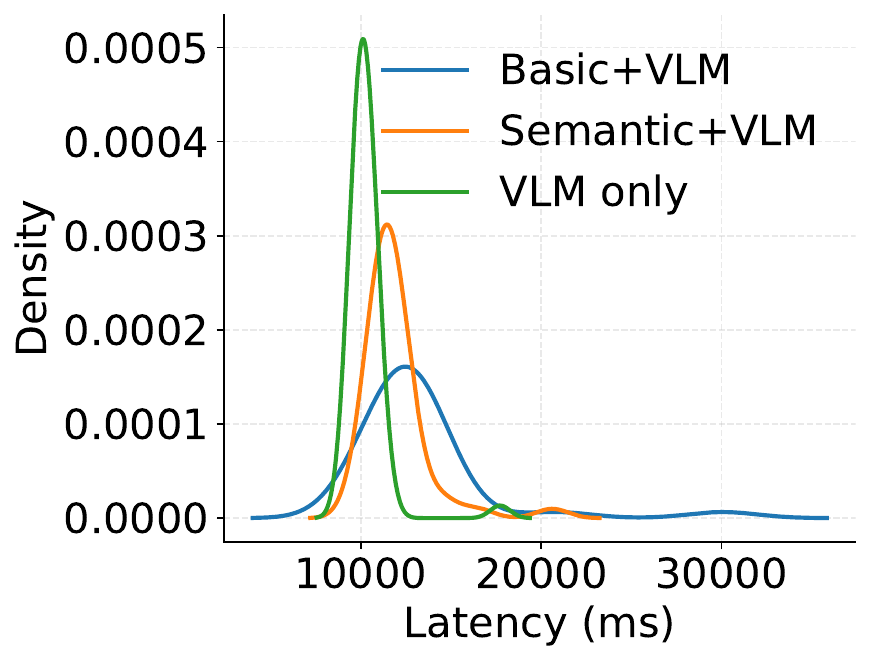}
        \caption{Latency PDF}
        \label{fig:vlm_latency_pdf}
    \end{subfigure}
    \begin{subfigure}[t]{0.24\linewidth}
        \centering
        \includegraphics[width=\linewidth]{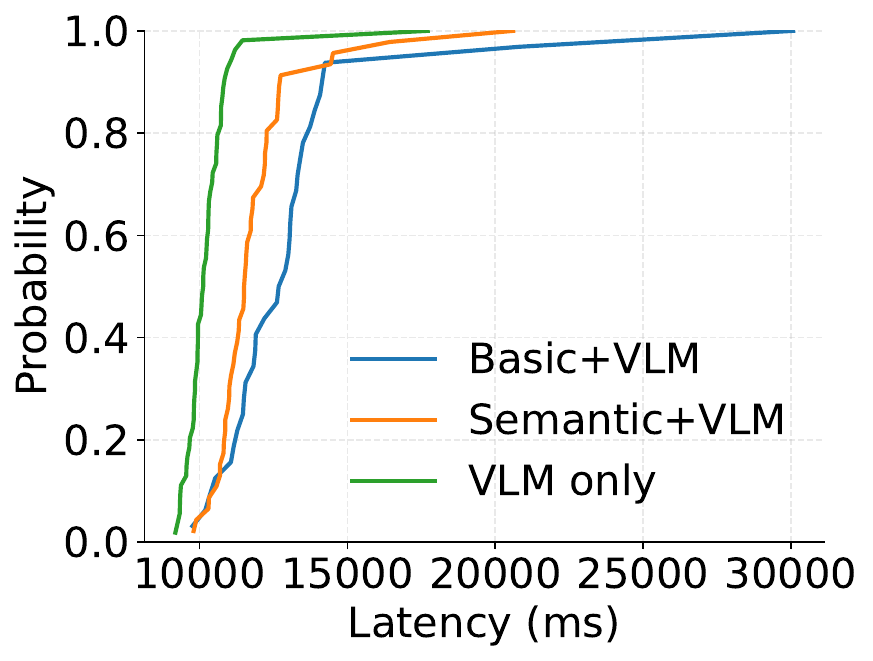}
        \caption{Latency CDF}
        \label{fig:vlm_latency_cdf}
    \end{subfigure}
    \caption{Distributions of VLM token generation throughput and end-to-end latency (Semantic + \texttt{gemma3:4b}).}
    \label{fig:vlm_distributions}
\end{figure*}

\begin{figure}[!t]
    \centering
    \includegraphics[width=\columnwidth]{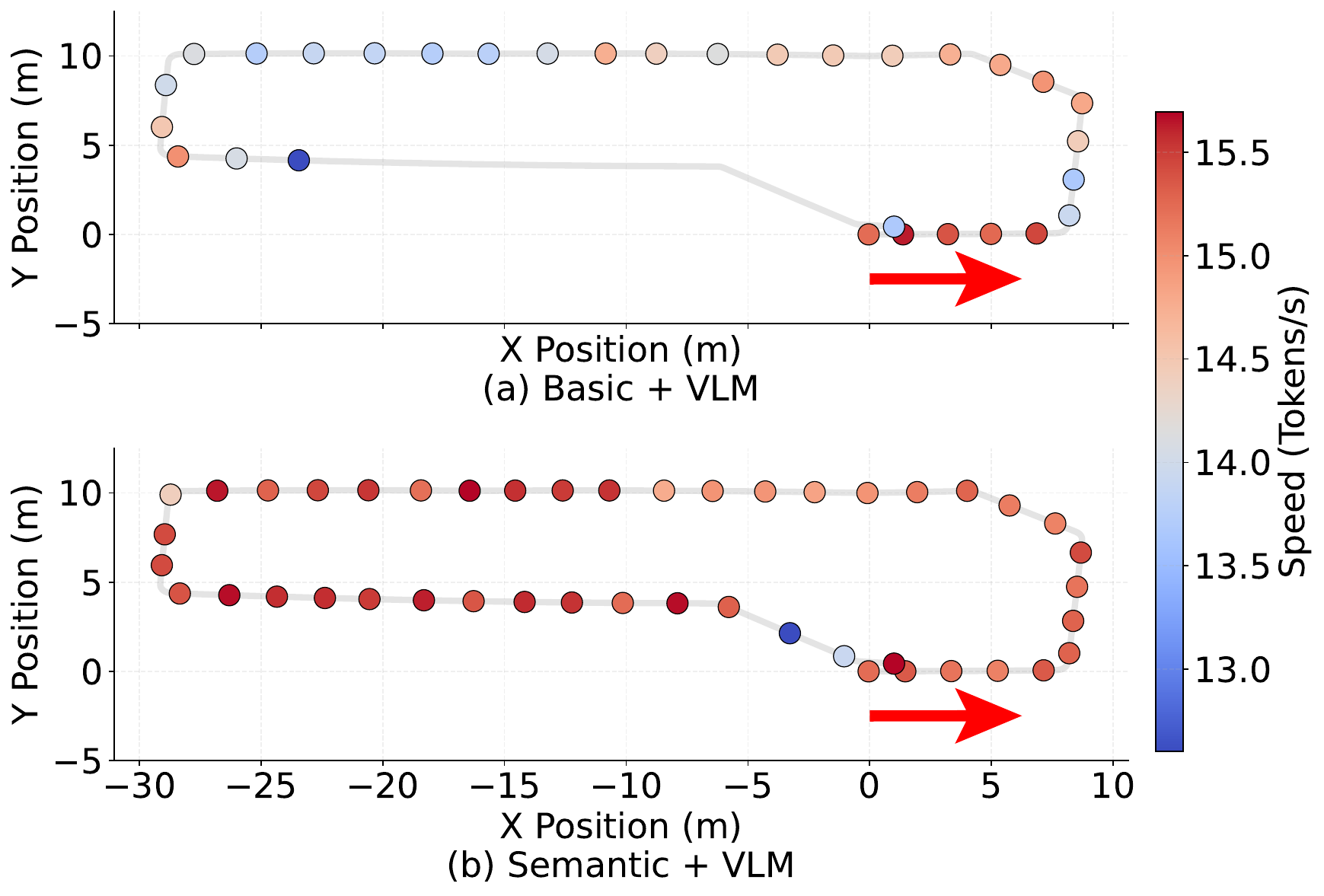}
    \caption{Token generation throughput with respect to robot's position during the experiment.}
    \label{fig:spatial_vlm_tps}
\end{figure}

\begin{figure}[t]
    \centering
    \includegraphics[width=\columnwidth]{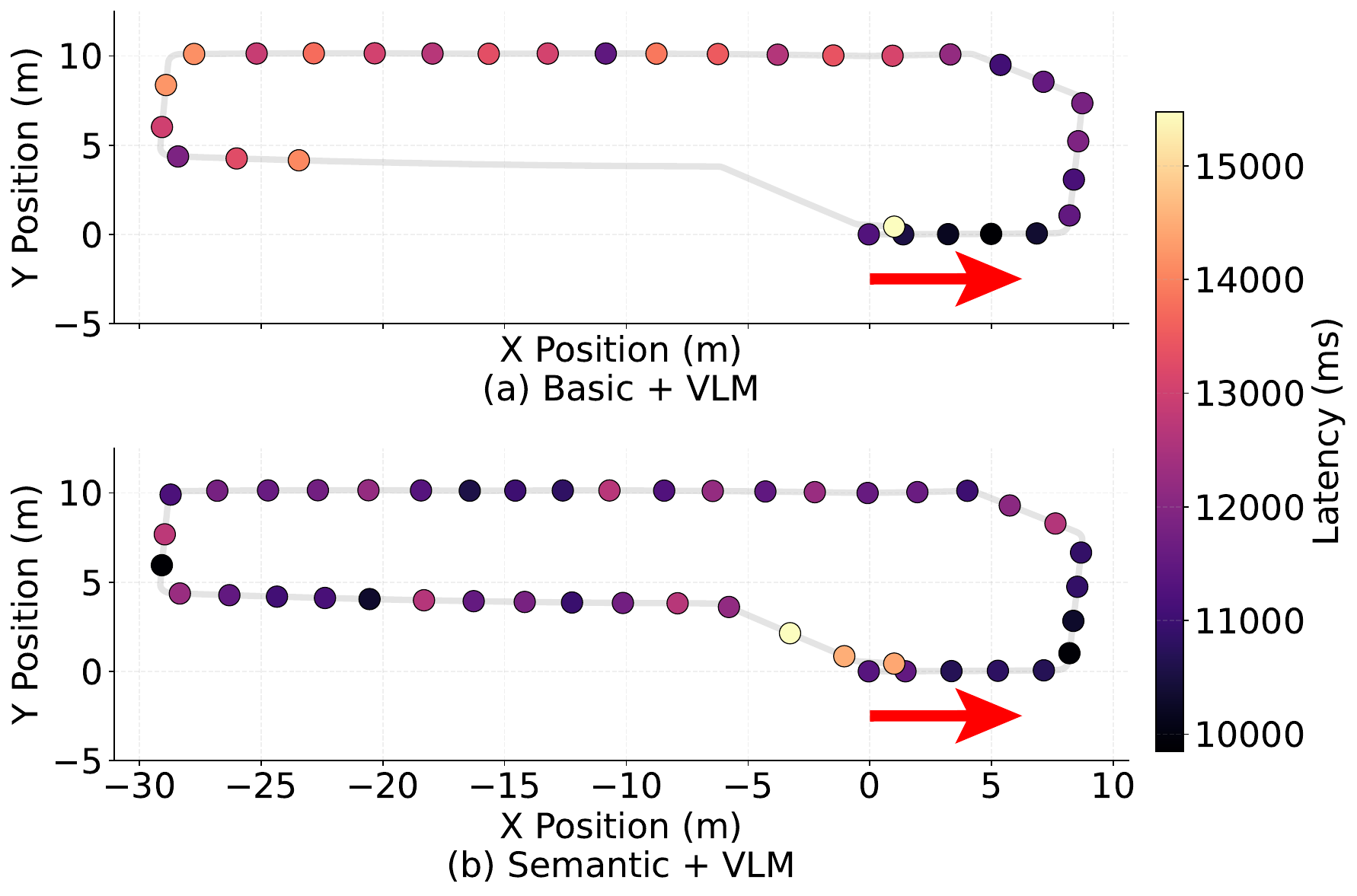}
    \caption{Token generation latency with respect to robot's position during the experiment.}
    \label{fig:spatial_vlm_latency}
\end{figure}

Figure~\ref{fig:experiment_trajectories} shows a trajectory comparison between GT, the baseline configuration, baseline+VLM, and semantic+VLM. The final approved loop closure occurs at the location indicated by the blue circle, and no further loop closures are approved until the scenario ends. This indicates that the baseline algorithm failed to properly perform loop closure due to insufficient memory, preventing keyframes from being loaded in time.

In contrast, the semantic zone based memory management exhibits much more stable performance under the same concurrent execution conditions. Enabling VLM only slightly increases the update rate from 0.96 Hz to 0.95 Hz, and the average total processing time increases from 566.06 ms to 784.81 ms. However, this increase is much smaller than the increase observed in the baseline algorithm, and the performance difference is also much smaller. Overall, these results show that semantic zone memory management supporting more robust dense-map utilization under concurrent edge-AI workloads.

\subsection{Impact of Concurrent SLAM-VLM Execution on VLM Performance}


To validate and evaluate the operational effectiveness of our proposed approach, we visualized the VLM inference metrics, Token Generation Speed and End-to-End Latency, spatially over the robot's physical path with Figure~\ref{fig:vlm_distributions}, \ref{fig:spatial_vlm_tps}, and \ref{fig:spatial_vlm_latency}.
A direct comparison between two configurations, Basic + VLM (baseline) and Semantic + VLM (proposed), reveals distinct pipeline characteristics.
The baseline configure ran VLM with the default SLAM configuration, which uses a conventional keyframe retrieval scheme.
And the other experiment, proposed, ran VLM with our semantic zone based approach, which acquires keyframes based on semantic zone information.
Among the two experiments, the VLM operates under identical inference conditions.

The baseline case frequently experiences severe latency spikes, which presented with annotated by the brighter, resulting in a token generation rate of 14.35 tokens/s with an average latency of 12.93 seconds.
These performance drops typically align with complex regions of the environment where dense, unfiltered visual data overwhelms the VLM's processing capacity.
Even worse, this approach fails to generate token after approximately 370 seconds of operation, then freeze for over 142 seconds until reach to the end point of the trajectory.

However, our proposed approach achieved robust operation and also presents better VLM metrics over baseline.
With our approach, VLM continuously generated tokens with the average speed of 15.19 tokens/s with an average latency of 11.68 seconds, without any freeze.
Based on these results, we were able to confirm that our proposed method could enables VLM to produce better results in terms of token generation speed, delay, and operational stability.

\section{DISCUSSION}

We proposed a semantic zone-based map management scheme and, which technique for efficiently utilizing memory in memory starving environments. We demonstrated that our proposed approach can simultaneously utilize memory-intensive RTAB-Map and VLM models with limited memory space. However, the following considerations must be considered.

During our experiments, we observed that when the memory limit is reached, additional keyframes cannot be loaded into memory from the map database. This restriction may delay or prevent keyframe retrieval required for loop closure, ultimately causing loop closure failure despite the availability of correct correspondences in the database. Such behavior highlights that memory saturation is not merely a performance issue but can directly affect localization stability in edge environments.

Several limitations remain. First, semantic zones were manually defined in the current study. Using an automated or adaptive spatial segmentation strategies may further improve scalability and reduce manual configuration effort. Second, the eviction policy was based on a least recently used (LRU) rule. More advanced strategies that incorporate task context, robot trajectory prediction, or semantic importance may yield improved keyframe selection efficiency. 

Future work may explore dynamic zone refinement, semantic importance weighting, and integration with online map updates. Additionally, more systematic evaluation of retrieval hit ratio and long-term deployment stability would further clarify the benefits of semantic aware keyframe selection under varying workload conditions.

\section{CONCLUSION}
We presented a semantic zone-based map management approach that stabilizes dense-map utilization under strict memory constraints. By associating keyframes with semantic zones and applying prediction-guided zone activation and eviction, the method enforces a fixed memory budget while preserving spatially relevant map content and reducing frequent memory accesses caused by keyframe loading and unloading. The zone-level batch operations make memory usage more predictable and prevent saturation that would otherwise delay keyframe retrieval and destabilize loop closure.

To validate our approach, we employed SLAM and VLM functionality on NVIDIA Jetson Orin Nano, which has 8 GB of its main memory space, then conduct experiment in Issac Sim based large-scale indoor environment. Throughout the experiment, our proposed method showed improved localization stability and loop-closure performance compared to the geometric strategy. Also, the proposed method showed better performance with respect to token generation throughput and latency of VLM models. With Qwen3.5:0.8b, the proposed method improves throughput by 3.3 tokens/s and reduces latency by 21.7\% relative to a geometric map-management strategy. Furthermore, while the geometric strategy suffers from out-of-memory failures and stalled execution under memory pressure, the proposed method eliminates both issues, preserving localization stability and enabling robust VLM operation.

These results indicate that the proposed approach could be a practical mechanism for deploying AI-integrated mobile robots on resource-constrained platforms. Beyond improving immediate runtime stability, our approach provides a systems-level strategy for balancing dense-map SLAM and VLM workloads within a limited memory budget. Overall, the proposed approach offers a concrete step toward reliable real-time robotic services that combine dense spatial mapping with large AI models on edge hardware.



\section*{Statement on Generative AI Use}
Generative AI tools were employed for language editing and grammar refinement and did not contribute to the research content of this paper.


\bibliography{references}

\end{document}